\documentclass[sigconf]{acmart}

\usepackage{multirow}
\usepackage{enumitem}
\AtBeginDocument{%
  }

\copyrightyear{2026}
\acmYear{2026}
\setcopyright{cc}
\setcctype{by}
\acmConference[RecSys '26]{20th ACM Conference on Recommender Systems}{September 27-October 02, 2026}{Minneapolis, MN, USA}
\acmBooktitle{20th ACM Conference on Recommender Systems (RecSys '26), September 27-October 02, 2026, Minneapolis, MN, USA}
\acmDOI{10.1145/3773078.3831787}
\acmISBN{979-8-4007-2284-4/2026/09}

\DeclareMathOperator*{\argmax}{arg\,max}

\begin{document}

\title[Inverse Theory of Mind Modeling for Content Recommendation]{Inverse Theory of Mind Modeling for Content Recommendation: From Web Browsing to Dynamic Intelligent Interfaces}

\author{Mengyu Chen}
\authornote{The first two authors contributed equally to this research.}
\email{mengyu.chen@jpmchase.com}
\affiliation{%
  \institution{JPMorganChase}
  \city{New York City}
  \state{NY}
  \country{USA}
}

\author{Feiyu Lu}
\authornotemark[1]
\email{feiyu.lu@jpmchase.com}
\affiliation{%
  \institution{JPMorganChase}
  \city{New York City}
  \state{NY}
  \country{USA}
}

\author{Chun-Fu (Richard) Chen}
\authornote{Corresponding authors.}
\email{richard.cf.chen@jpmchase.com}
\affiliation{%
  \institution{JPMorganChase}
  \city{New York City}
  \state{NY}
  \country{USA}
}

\author{Lucas Vinh Tran}
\authornotemark[2]
\email{lucas.vinhtran@chase.com}
\affiliation{%
  \institution{JPMorganChase}
  \city{London}
  \country{UK}
}

\author{Jay Katukuri}
\email{jay.katukuri@chase.com}
\affiliation{%
  \institution{JPMorganChase}
  \city{Palo Alto}
  \state{CA}
  \country{USA}
}

\renewcommand{\shortauthors}{Chen et al.}
\newcommand{\rc}[1]{{\color{magenta}{\small\bf\sf [RC: #1]}}}
\newcommand{\fl}[1]{{\color{aquamarine}{\small\bf\sf [FL: #1]}}}
\newcommand{\tablevspace}{{\vspace{-6pt}}}
\begin{abstract}
    Modern recommender systems treat observed actions as reliable proxies for user
preferences, yet interactions often reflect exploration or comparison rather
than stable preference expression.
As interfaces evolve from static layouts toward generative UIs and immersive
extended reality~(XR), the need for deeper, modality-agnostic user
understanding grows: these adaptive environments must decide not only
\emph{what} to present but \emph{where}, \emph{when}, \emph{how prominently},
and most importantly \emph{why} a user acts.
We propose an Inverse Theory of Mind (IToM) pipeline that reasons backward
from observed interactions to infer the beliefs, preferences, and
decision-making traits that explain behavior.
The pipeline reconstructs each user's decision context, including what was
chosen and what alternatives were available, applies LLM-driven counterfactual
reasoning
to produce evidence-grounded natural-language belief statements, and
synthesizes these beliefs through multi-hypothesis abductive inference into a
structured \emph{user persona}.
We evaluate on the OPeRA dataset against ground-truth personality assessments,
attitudinal surveys, and interview-based personas across four tasks: next
action prediction, shopping attitude alignment, Big Five personality
inference, and held-out category prediction.
Results show that inferred personas match or exceed ground-truth personas and
that multi-hypothesis reasoning is essential for accurate personality
prediction.
We further demonstrate cross-modal transferability with a persona-driven spatial
banking application on VisionOS.
    \end{abstract}

\ccsdesc[500]{Information systems~Recommender systems}
\ccsdesc[300]{Human-centered computing~Mixed / augmented reality}
\ccsdesc[500]{Computing methodologies~Reasoning about belief and knowledge}
\ccsdesc[500]{Computing methodologies~Natural language generation}
\keywords{Theory of Mind, User modeling, Persona synthesis, Recommender systems, LLM reasoning, Generative UI, Extended reality, Adaptive interfaces,  Persona prediction}


\maketitle

\section{Introduction}
Recommender systems have made substantial progress in predicting user behavior
from large-scale interaction histories, including clicking, viewing, and scrolling.
Methods such as collaborative filtering~\cite{koren2009matrix,he2017neural},
sequential models~\cite{kang2018self,sun2019bert4rec}, and generative retrieval
with semantic item IDs~\cite{rajput2024recommender} recommend items by learning
latent patterns from past interactions.
More recently, Large Language Model (LLM)-based recommenders have emerged,
using language understanding and reasoning for more conversational and
explainable recommendations~\cite{geng2022recommendation,wu2024survey}.

Despite these advances, there is limitation with such methods.
\emph{Interaction based} systems learn statistical patterns over a specific
behavior type (e.g., clicks, purchases, ratings) and can only predict
the next behavior of that same type~\cite{lo2016understanding,guo2025longer}.
Even recent approaches that scale sequence length~\cite{guo2025longer} or
integrate collaborative semantics into LLMs~\cite{zheng2024adapting} remain
bound to this scope.
At the same time, user interfaces are evolving toward generative
UIs~\cite{swearngin2018rewire,li2024personal} that compose novel layouts on the
fly, which aims at offering personalized experiences that adapt to particular users and devices (e.g., mobile, desktop, and spatial extended reality (XR) systems).
These emerging paradigms need to decide not only \emph{what} to show but
\emph{where}, \emph{when}, and \emph{how} to present it, which demands a
richer model of the user than click histories can provide.
\emph{Persona-based} approaches offer that richer model by constructing
profiles that capture interests, goals, personality traits, values, and life
context~\cite{zhang2024generative,park2023generative}, but they traditionally
require qualitative data such as user interviews, demographic surveys, or
manually authored persona
templates~\cite{chen2025personatwin,li2025digital}, which are expensive,
intrusive, and difficult to scale.

The common missing piece is an understanding of the \emph{rationale} behind user behavior.
In cognitive science, reasoning about others' beliefs, goals, and intentions is known as Theory of Mind
(ToM)~\cite{premack1978does,wellman2014making}.
ToM is well suited to user modeling because it is context-dependent,
interpretable, and modality-agnostic.
By applying ToM \emph{inversely}, reasoning backward from observed actions to
the beliefs that generated them, we can extract persona-level understanding
from ordinary interaction traces without requiring qualitative data.

Building on this perspective, we propose an \emph{Inverse Theory of Mind}
(IToM) pipeline that performs \emph{abductive inference}, reasoning from
observed behavior to its best explanation, over user interaction traces.
Our pipeline transforms raw event metadata into
human-readable belief representations through action summarization, context
reconstruction, and optimization. We then synthesize these
user beliefs into a structured \emph{user persona}: a cognitive profile capturing
the user's interests, motivational drives, life context, and values.
We use LLM as the reasoning engine at each stage, producing evidence-grounded
belief statements and persona profiles rather than opaque latent vectors.
Once constructed, the persona can be used as a portable user
model for tasks such as personality-aligned attitude inference, shopping preference
forecasting, or content adaptation for novel interfaces. Applications can use the persona
through standard prompting without task-specific
retraining~\cite{zhang2024generative}.

We validate the approach on the OPeRA dataset~\cite{wang2025opera}, which pairs
fine-grained user action logs from Amazon shopping sessions with ground-truth
interview transcripts, personality assessments, and shopping attitude surveys.

Our contributions are:
\begin{itemize}[leftmargin=*, noitemsep, topsep=2pt]
       \item We propose \emph{Inverse Theory of Mind} as a paradigm that bridges
    interaction-based recommenders (data-rich but scope-bound) and persona-based
    methods (rich understanding but dependent on costly qualitative data), and
    instantiate it as a multi-stage abductive inference pipeline that
    reconstructs decision contexts, infers counterfactual-grounded beliefs, and
    synthesizes them into a modality-agnostic user persona from behavioral
    traces alone.

    \item We introduce \emph{multi-hypothesis persona synthesis}, generating
    diverse personality interpretations of the same behavioral evidence and
    aggregating them with adaptive confidence weighting.

    \item We validate on real-world browsing data (OPeRA), comparing inferred
    personas against ground-truth interviews and personality assessments,
    demonstrating that behaviorally inferred personas support downstream tasks
    (shopping-attitude alignment, personality inference, held-out category prediction) beyond the scope
    of interaction-based methods.

    \item We present an illustrative spatial banking prototype on VisionOS as a proof-of-concept for persona portability across interaction modalities, where personas inferred from 2D browsing data inform 3D layout and content selection.

\end{itemize}

\section{Related Work}
\subsection{User Modeling and Intent Inference}

Recommender systems have progressed from matrix factorization on explicit
ratings~\cite{koren2009matrix} through implicit-feedback
methods~\cite{hu2008collaborative,rendle2009bpr,TranT0CL20} to deep sequential models
that capture temporal patterns via recurrent networks, self-attention, bidirectional
transformers and beyond~\cite{hidasi2016session,kang2018self,sun2019bert4rec,FengTCCLL20}.
Recent generative retrieval and industrial-scale systems push this paradigm
further by framing recommendation as token generation over learned item
IDs~\cite{rajput2024recommender} or modeling tens of thousands of
interactions~\cite{guo2025longer}.
Integrating LLMs into recommendation has opened additional directions,
including text-to-text task unification~\cite{geng2022recommendation}, LLM
fine-tuning and instruction-following for
recommendation~\cite{bao2023tallrec,zhang2023instructrec}, zero-shot
ranking and conversational
recommendation~\cite{hou2024large,he2023zeroshot}, prompt enrichment
strategies for personalized
recommendation~\cite{lyu2024llmrec}, and hybrid collaborative-language
reasoning~\cite{zhu2024collaborative,zheng2024adapting}.
Recent work explores LLM-driven user interest profiling from interaction histories~\cite{christakopoulou2023interest,mysore2024pearl,richardson2024userllm}. PEARL~\cite{mysore2024pearl} retrieves and re-ranks representative snippets from a user's existing text; \citet{christakopoulou2023interest} cluster consumption logs into LLM-generated interest journeys; and UserLLM~\cite{richardson2024userllm} conditions generation on precomputed embedding-based user representations.
A parallel line argues for natural-language user profiles as a route to transparent and scrutable recommendation~\cite{radlinski2022natural,ramos2024transparent}, typically assembling profiles from explicit user statements or curated text.
Complementing these, session-based approaches model intra-session dynamics
through graph neural networks or attention
mechanisms~\cite{wu2019session,li2017neural}, e-commerce models classify
browsing versus purchasing
intent~\cite{lo2016understanding,guo2019buying}, and probabilistic click
models formalize examination and click decisions in web
search~\cite{chuklin2015click}.
In planning and robotics, goal recognition treats intent inference as an
inverse problem over observed action
sequences~\cite{ramirez2010probabilistic,sukthankar2014plan}.

These methods, including LLM-enhanced variants, achieve strong next-item or
next-action prediction yet remain scope-bound: they
model statistical regularities within a single interaction type, reduce intent
to a categorical label or scalar reward, and cannot transfer to tasks
requiring a different domain of user understanding, such as personal goals,
lifestyle profiling, or adaptation to an unseen interface modality.

Inverse reinforcement learning (IRL) moves closer to latent motivations by
recovering reward functions from demonstrated
trajectories~\cite{peysakhovich2019irl,chen2020inverse}.
Zhao et al.~\cite{zhao2025mtrec} introduced mental reward models that
use distributional IRL to infer users' internal satisfaction, and
Wang et al.~\cite{wang2022causal} discover causal rather than co-occurrence
relations among user actions, improving both accuracy and explainability.
These approaches produce richer signals than categorical intent labels but
still operate in numeric reward spaces rather than interpretable
representations of why a user acts.

A separate line of work constructs user personas to drive
recommendation and
simulation~\cite{zhang2024generative,park2023generative,zhang2025see}.
However, these approaches assume that the persona is given as input,
authored manually or derived from qualitative sources such as demographic and
personality
profiles~\cite{chen2025personatwin,li2025digital}, rather than
inferred from behavioral traces.
Our work bridges these two directions by introducing belief inference as an
intermediate representation, constructing personas from behavioral traces without qualitative data.

\subsection{Theory of Mind in AI}

Theory of Mind (ToM) refers to the cognitive ability to attribute beliefs,
desires, and intentions to
others~\cite{premack1978does,baroncohen1985does}.
Computational approaches formalize ToM as Bayesian inverse planning over goals
and
beliefs~\cite{baker2009action,baker2017rational,jaraettinger2019theory,mao2024review},
and neural meta-learning has shown that ToM-like reasoning can be acquired from
agent trajectories~\cite{rabinowitz2018machine}.
Recent multi-agent work applies ToM to LLM-based collaboration through explicit
belief tracking, hypothesis-driven goal attribution, and inverse reinforcement
learning~\cite{li2023theory,cross2024hypothetical,wu2023multiagent}, though
these remain confined to game-like or synthetic settings.

Evaluations confirm that LLMs can pass false-belief tasks and perform
higher-order belief
reasoning~\cite{kosinski2024evaluating,street2024llms}, though with notable
gaps on certain
tasks~\cite{sap2022neural,ullman2023large,shapira2024clever,strachan2024testing,chen2024tombench}.
ToM-inspired reasoning has also entered applied settings recently.
Liu et al.~\cite{liu2025mindful} show that ToM-aware recommendations increase
consumer trust, and Pawar et al.~\cite{pawar2025earl} apply ToM to
surface user intent from partial interaction sequences.
These results indicate that modeling why a user acts yields richer
understanding than predicting what they do next, yet all operate in
forward-reasoning or single-intent settings.

\subsection{Personalization for Adaptive and Spatial Interfaces}

Emerging interface paradigms, including generative UIs that compose layout and
content dynamically~\cite{swearngin2018rewire,li2024personal} and
XR
environments that distribute content across the user's visual
field~\cite{grubert2018challenges}, require the system to decide not only
what to show but where, when, and how prominently,
demanding richer user models than click histories alone can provide.
Current XR personalization relies predominantly on in-session physiological and
attentional signals such as gaze, head pose, RL-driven placement, and
affective
biometrics~\cite{clay2019eye,lindlbauer2019context,lu2024adaptive,marchiori2018narratives}.
Li et al.~\cite{li2025satori} move toward deeper modeling by applying
Belief-Desire-Intention reasoning to proactive AR assistance, though their BDI
states are hand-designed rather than inferred from data.
These methods are effective within a session but face severe cold-start
limitations.
A complementary strategy bootstraps personalization from data collected in
established modalities; however, existing LLM-based persona
approaches~\cite{park2023generative,li2025digital,chen2025personatwin}
construct personas from qualitative sources such as surveys or comments, rather than \emph{inferring} them from behavioral traces.

\section{Methodology}

\subsection{ToM-Inspired User Modeling}
\label{sec:tom_modeling}

Bridging the gap between interaction-based and persona-based methods requires extracting the \emph{rationale} behind each action.

This relies on \emph{abductive belief inference} grounded in Theory of
Mind~(ToM).
Rather than reasoning forward from beliefs to predict actions, we reason
\emph{backward} from observed actions to infer the beliefs that generated
them, a form of inference to the best
explanation~\cite{baker2009action,jaraettinger2019theory}.
Three properties of ToM make it particularly suited to this task.

\emph{Context-dependence.}
ToM infers mental states from the relationship between an action and the
alternatives available: a choice is only meaningful relative to what was
\emph{not} chosen, so reconstructing the decision context could help us extract the rationale behind an action.

\emph{Semantic Structure}
ToM decomposes user behavior into semantically meaningful primitives (beliefs, desires, and intentions) rather than a single opaque representation. These primitives are compositional (a user model is built up from many small, attributable statements) and directly refer to mental states that could explain specific actions, giving downstream inference a structure that latent-vector representations do not provide. When serialized in natural language, as in our LLM instantiation, this representation additionally supports interpretable and auditable user profiles.

\emph{Modality-agnosticism.}
Because beliefs describe \emph{why} a user acts rather than encoding
interface-specific patterns, they transfer across modalities without retraining.
A belief inferred from screen-based browsing is equally actionable for a
generative UI engine composing a personalized layout or a spatial XR interface
arranging content in the user's visual field.

\emph{From beliefs to personas.}
While individual beliefs capture micro-level decision rationales, they can be
further synthesized into a structured \emph{persona}, a portable cognitive profile of the user's interests, values, and decision-making style.
This synthesis performs a second layer of abductive inference, moving from
``what did the user prefer?'' to ``what kind of person makes these choices?''.
The resulting persona could enable not only preference prediction but also explanation of past choices and adaptation to novel contexts where no interaction history exists.

Given these properties, we decompose user understanding into two explicit sub-problems: (i)~\emph{percept reconstruction} of the decision context, and (ii)~\emph{abductive belief inference} from the observed choice within that context, followed by (iii)~\emph{persona synthesis} that integrates atomic beliefs into a coherent cognitive profile. This inverse reasoning requires a flexible engine capable of processing unstructured interface content and performing causal, counterfactual judgments about human behavior, a form of everyday causal reasoning that LLMs have shown increasing competence in~\cite{joshi2024cold,strachan2024testing}. 
We therefore use LLMs not as recommenders but as approximate inverse ToM reasoners, guided by structured prompts to produce evidence-grounded and human-readable belief statements.
LLM reasoning is nevertheless known to regress toward stereotypical priors and to vary across paraphrased prompts~\cite{sap2022neural,ullman2023large,shapira2024clever,strachan2024testing,chen2024tombench}; our pipeline mitigates these failure modes by two principles that recur across stages: 1) grounding each inference in a reconstructed choice context rather than free-form recall, and 2) aggregating over multiple competing interpretations rather than committing to a single one.

\subsection{Problem Formulation}

Let $T_u = \{S_1, \ldots, S_K\}$ denote the interaction trace of user~$u$ across $K$ sessions. Each session $S_k = (a_1^k, \ldots, a_{n_k}^k)$ is a time-ordered sequence of interface actions, where each action
\[
  a_i^k = (\tau_i,\; \mathit{target}_i,\; \mathit{ctx}_i,\; t_i),
\]
consists of an action type $\tau_i \in \{\texttt{click},\, \texttt{input},\, \texttt{scroll},\, \texttt{nav},\, \texttt{tab}\}$, a target element, contextual metadata, and a
timestamp. While we instantiate the pipeline on screen-based web interactions, the formulation generalizes to any modality that produces event-level action traces with associated interface context (e.g., mouse clicks, finger taps, gaze selections and gesture inputs in
XR, or component-level interactions in generative UIs).

Our objective is to infer a belief set
$\mathcal{B}_u = \{b_1, \ldots, b_m\}$ where each $b_j$ is a natural language
statement capturing a preference, trait, or intention, and to synthesize these
beliefs into a structured persona~$\Pi_u$.
We state the \emph{conceptual objective} as follows:
\begin{equation}
  \mathcal{B}_u^* \;=\; \argmax_{\mathcal{B}}\; P(\mathcal{B} \mid T_u)
\end{equation}
under a rational-agent assumption: observed actions are explainable by the
user's beliefs given the decision context they perceived.
The persona is then obtained as
\begin{equation}
  \Pi_u \;=\; \mathrm{Synthesize}(\mathcal{B}_u^*)
\end{equation}
where $\mathrm{Synthesize}$ performs a second round of abductive inference,
integrating atomic beliefs into a coherent cognitive profile of the user's
interests, values, and decision-making style.

Given the persona $\Pi_u$, downstream prediction reduces to conditional
inference: for a task-specific query $q$ (e.g., ``rank these products by
preference'' or ``predict Big Five trait scores''), the system computes
\begin{equation}
  \hat{y} \;=\; \argmax_{y}\; P(y \mid q,\, \Pi_u)
\end{equation}
where the persona provides the contextual grounding that enables the model to
reason about what the user would do, how they would think, and what they would
prefer in a given situation, without requiring additional interaction data
for each new task.

\paragraph{Approximation strategy.}
Equations~1--3 state inferential \emph{objectives}; exact posterior
computation over the space of natural-language beliefs is intractable.
We instead adopt an \emph{amortized approximate inference}
strategy~\cite{gershman2014amortized,stuhlmuller2013learning}:
an LLM pre-trained on large-scale text has internalized rich priors over
human beliefs and decision-making
patterns~\cite{binz2023turning,strachan2024testing}; conditioned on
a reconstructed decision context through structured prompts, it acts as
an approximate inference engine that produces candidate beliefs without
an explicit likelihood function or posterior computation.
Eq.~1 can thus be approximated by conditioning the LLM on each decision
context in the trace to generate candidate beliefs, which can then be
refined for diversity and coverage.
Eq.~2 can be approximated by prompting the LLM to synthesize the refined
beliefs into a coherent persona, and Eq.~3 can be realized by conditioning
the LLM on the resulting persona for task-specific queries.
The persona thus acts as an \emph{approximate sufficient representation}:
a lossy but task-relevant compression of the full trace that decouples
prediction from raw interaction data.

\subsection{Pipeline Overview}

The pipeline transforms raw interaction events into a structured user persona through five sequential stages:

\begin{enumerate}[leftmargin=*, noitemsep, topsep=4pt, label=\textbf{(\arabic*)}]
  \item \textbf{Summarize:} Convert raw data into natural language sentences.
  \item \textbf{Perceive:} Reconstruct the decision context, revealing what alternatives were visible on the page.
  \item \textbf{Infer:} Perform backward belief inference at different levels.
  \item \textbf{Optimize:} Deduplicate and select a diverse, compact belief set.
  \item \textbf{Synthesize:} Compose beliefs into a structured persona.
\end{enumerate}

\subsection{Stage 1: Action Summarization}

Raw interaction events contain type-specific fields that are not directly interpretable.
This stage converts each action into a concise natural language summary, for example:

\smallskip
\noindent\fbox{\parbox{0.95\columnwidth}{\small
\textbf{Input:} \texttt{click} on ``Cedar Wood (Square Ceramic)'' color variant \\
\textbf{Output:} ``\emph{User selected the `Cedar Wood (Square Ceramic)' color option.}''
}}
\smallskip

\noindent Structurally simple actions (scroll, navigation, tab activation) are summarized by deterministic rules, while complex actions (click, input) are summarized by an LLM that incorporates product details when present. 

\subsection{Stage 2: Percept Reconstruction}
\label{sec:perception}

This stage addresses a key limitation of standard behavioral analysis: actions are recorded without the decision context in which they occurred. A user's selection is only informative relative to the alternatives they could perceive. We reconstruct this context by analyzing the page content alongside the action
summary.

For each action, we prompt an LLM with the action summary, a condensed version of the page HTML, and page metadata. The LLM produces a structured \emph{percept} containing:

\begin{itemize}[leftmargin=*, noitemsep, topsep=2pt]
    \item \textbf{Main choices}: Options highly relevant to the user's action, representing the primary decision set.
    \item \textbf{Sub choices}: Secondary alternatives that could have been picked.
    \item \textbf{Other choices}: Available but irrelevant options on the page.
    \item \textbf{User choice}: What the user actually did.
    \item \textbf{Perception description}: A summary of the overall page environment and what the user likely encountered.
\end{itemize}

The three-tier choice classification is central to our belief inference: main choices represent the ``consideration set'' from which the user selected, sub choices define the broader option space, and other choices provide contrast for what the user did \emph{not} pursue. Together, they form the decision context needed for inverse reasoning.

\smallskip
\noindent\fbox{\parbox{0.95\columnwidth}{\small
\textbf{Example percept} (cart page): 
\textbf{Main:} Air freshener variants, checkout.
\textbf{Sub:} Quantity adjustment, Subscribe \& Save.
\textbf{Other:} Unrelated cart items.
\textbf{User chose:} Ocean Mist variant --- ``\emph{view details for item in cart}.''
}}
\smallskip

\subsection{Stage 3: Backward Belief Inference}

We consider this the core contribution of our work, which, to the best of our knowledge, little prior research has explored for personalization and recommendations. Given the percept reconstruction from Stage~2 (what the user saw, what they chose, what alternatives existed), we perform \emph{backward} reasoning to infer the beliefs that best explain the observed behavior.
We operate at two complementary levels.

\subsubsection{Action-Level Beliefs.}
For each individual action, the LLM receives the complete percept, including
action summary, user choice with rationale, all three tiers of available
choices, and the perception description, and generates multiple
candidate belief inferences (1 sentence each), representing different plausible interpretations of the same observed behavior.

\smallskip
\noindent\fbox{\parbox{0.95\columnwidth}{\small
\textbf{Percept:} Selected ``Cashmere Vanilla (Tassel Ceramic)'' among 8 scent/design variants. \\
\textbf{Belief:} ``\emph{Values subtle luxury and aesthetic harmony, choosing a decorative variant over simpler options.}''
}}
\smallskip

Action-level beliefs capture \emph{micro-signals}: specific product preferences, fine-grained decision-making traits, and individual action rationales.

\subsubsection{Session-Level Beliefs.}
After processing all actions in a session, the LLM receives the full sequence of
perceived items together with the action-level belief hypotheses in temporal order and generates multiple candidate belief summaries (each 1-2 sentences) covering the session as a whole. Building upon the action-level beliefs, the session-level prompt treats them as competing candidate explanations and identifies which combination best fits the overall pattern.

\smallskip
\noindent\fbox{\parbox{0.95\columnwidth}{\small
\textbf{Session:} 33 actions --- searched for a specific bicycle tail light, compared handlebar tape colors/specs, read reviews. \\
\textbf{Belief:} ``\emph{Dedicated cyclist who prioritizes safety; research-driven purchasing style with methodical comparison before selection.}''
}}
\smallskip

Session-level beliefs capture \emph{macro-patterns}: cross-action consistency, temporal dynamics (e.g., exploration narrowing into decision), lifestyle signals (e.g., product category mix), and overall decision-making style. As session-level inference processes dozens of actions at once and may dilute fine-grained signals from individual decisions, action-level beliefs preserve those micro-signals which help capture the full picture. 

\subsubsection{Grounding and belief faithfulness.}
Hallucination is constrained by construction that every belief is grounded in a reconstructed percept (Stage~2), must cite the chosen option and a rejected alternative from that percept (Stage~3), and is aggregated across $M$ competing hypotheses (Stage~5) so that single-hypothesis fabrications are diluted rather than propagated.

\subsection{Stage 4: Belief Optimization}
\label{sec:optimization}

Stages~1--3 produce a candidate belief set that is heavily redundant
(337--3{,}167 candidates in our evaluation), with common, repetitive shopping-action paraphrases.
We select a compact, diverse subset via Maximal Marginal
Relevance~(MMR)~\cite{carbonell1998mmr} in two phases.

\textit{Phase 1: Deduplication and quality scoring.}
Beliefs are embedded with a dense text encoder; near-duplicate pairs
(cosine similarity~$> \theta_{\mathrm{dedup}}$) are collapsed, preferring
session-level beliefs.
Each survivor receives a quality score~$q(b)$ based primarily on its
\emph{outlier distance} from the embedding centroid, so that rare and distinctive beliefs score high and frequent shopping-action paraphrases score low.

\textit{Phase 2: MMR iterative selection.}
Starting from the highest-scored belief, we iteratively select up to $K$
beliefs by maximizing:

\begin{equation}
  \mathrm{MMR}(b) \;=\; \lambda \cdot q(b) \;+\;
    (1 - \lambda)\Bigl(1 - \max_{b' \in \mathcal{S}}\;
    \cos(b, b')\Bigr)
  \label{eq:mmr}
\end{equation}
where $\mathcal{S}$ is the already-selected set and
$\lambda = 0.3$ weights diversity at 70\%, ensuring that once a topic is
covered, remaining variants are penalized.
The output is a compact set
$\mathcal{B}_u^* = \{b_1, \ldots, b_K\}$ covering the user's full
behavioral range with minimal redundancy.

\subsection{Stage 5: Multi-Hypothesis Persona Synthesis}
\label{sec:persona}

Atomic beliefs describe individual episodes; downstream tasks require a holistic view of the user.
However, the mapping from browsing behavior to personality is
\emph{under-determined}, that the same pattern can arise from different personality types, and committing to a single interpretation risks defaulting to stereotypes.

Stage~5 therefore generates $M$ diverse \emph{persona hypotheses}, each a
distinct personality type that plausibly explains all observed behaviors.
The LLM receives the optimized belief set and is prompted to
(i)~separate observable facts from personality interpretations and
(ii)~consider whether the opposite trait extreme could equally explain the
evidence.
Each hypothesis produces:

\begin{itemize}[leftmargin=*, noitemsep, topsep=2pt]
    \item A \textbf{persona profile} with interests, drives, values, and life
    context, all inferred abductively from the beliefs.
    \item A structured \textbf{shopping profile} listing only product categories with direct evidence from the beliefs.
    \item Per-trait feature for personality, reflecting
    how well the behavioral evidence constrains each trait.
    \item \textbf{evidence confidence levels} (strong, moderate, or weak), reflecting how well the behavioral evidence constrains each trait.
\end{itemize}

\noindent Hypotheses are ranked by quality and distinctiveness, and combined
via confidence-weighted aggregation.
For any scalar dimension~$d$ (e.g., a Big Five trait or a Likert
attitude item), each hypothesis~$h$ independently produces a
score~$s_{h,d}$; the final prediction is:
\begin{equation}
  \hat{s}_{d} = \frac{\sum_{h=1}^{M} w_{h} \cdot s_{h,d} + \pi_{d} \cdot \alpha_{d}}{\sum_{h=1}^{M} w_{h} + \alpha_{d}},
  \label{eq:mhi}
\end{equation}
where $w_h$ combines hypothesis quality with per-dimension evidence
confidence, $\pi_{d}$~is a published population prior for dimension~$d$,
and $\alpha_{d}$~is an adaptive prior strength that scales inversely with
mean evidence confidence across hypotheses.
When evidence is strong the prior vanishes and the hypotheses speak;
when weak, predictions regress toward population means, mitigating the risk of overconfident stereotyping.

\section{Experiments}
\label{sec:experiments}

We evaluate the IToM pipeline on four complementary tasks, each posed as a research question (RQ) that isolates a different component of the pipeline:
(1)~next action prediction, testing whether inferred personas improve
behavioral simulation compared to ground-truth personas;
(2)~shopping category prediction, testing downstream predictive power on
held-out browsing sessions;
(3)~Big Five personality prediction, testing whether behavioral traces contain
recoverable personality signal; and (4)~shopping preference survey alignment, testing whether inferred personas
capture self-reported shopping attitudes. 
RQ1 tests whether the synthesized persona (Stage~5) provides usable behavioral signal downstream; RQ2 tests generalization of the persona to unseen sessions; RQ3 (with its \S4.5 ablation) tests whether multi-hypothesis aggregation matters; and RQ4 tests persona alignment with self-reported attitudes. 

\subsection{Dataset}

As our method essentially builds upon a quantitative-to-qualitative inference process, it imposes three simultaneous
requirements on the evaluation data:
(i)~\emph{fine-grained interaction traces} (clicks, scrolls, typed queries),
not just item-level ratings;
(ii)~\emph{rich decision context} (page content, visible alternatives) to
reconstruct what the user perceived when acting; and
(iii)~\emph{qualitative ground truth} (personality assessments, attitudinal
surveys, interview transcripts) to verify the inferred qualitative output.
Most existing benchmarks satisfy at most one of these.
Widely used recommender datasets such as
MIND~\cite{wu2020mind} and KuaiRec~\cite{gao2022kuairec}
provide rich interaction logs but no psychological or attitudinal ground truth.
Conversely, personality-oriented corpora such as
PANDORA~\cite{gjurkovic2021pandora} supply trait labels linked to text
but not the atomic behavioral traces from which beliefs can be inferred.
To our knowledge, no benchmark prior to OPeRA jointly provides all three.

We therefore evaluate on the OPeRA dataset~\cite{wang2025opera}, which
jointly provides all three types of data: atomic-level browser interaction traces with full page context (HTML snapshots) from Amazon shopping sessions, paired with
per-user personality assessments, shopping attitude surveys, and post-session interview transcripts.
We use the OPeRA \emph{train} split of 46 complete user profiles, personality and shopping surveys.

\subsection{Implementation Details}

All LLM calls use Claude Opus 4.6 via AWS Bedrock.
Text embeddings use Amazon Titan Embed v2 (1024 dimensions).
For belief optimization, we use MMR with $\lambda = 0.3$ and deduplication
threshold $\theta_{\mathrm{dedup}} = 0.85$, selecting up to $K = 50$ beliefs per user.
Multi-hypothesis persona generation uses $M = 7$ hypotheses with temperature 0.85. For personality and attitude inference, responses are predicted
using the confidence-weighted aggregation from
Eq.~\ref{eq:mhi} (Section~\ref{sec:persona}), with population priors set to published Big Five norms.

Per-user persona construction (mean $6.3$ sessions, $\sim$275 actions) requires $\sim$622 LLM calls and $\sim$14.5M input / $\sim$240K output tokens, amortizing to $\sim$\$79 per user on Claude Opus~4.6 ($\sim$\$47 on Sonnet~3.7) at AWS Bedrock pricing (as of April 2026). Construction is one-time; the resulting natural-language persona is reused across all downstream tasks with no per-query re-inference. The pipeline is an unoptimized research implementation where Stage~2 (percept reconstruction) alone accounts for $\sim$96\% of input tokens, with no batching of same-page actions, no cross-user caching of layout templates, and no per-action call consolidation.

\subsection{RQ1: Next Action Prediction}
\label{sec:rq1}

\paragraph{Setup.}
We evaluate whether inferred personas can substitute for ground-truth personas
in the next action prediction task introduced by OPeRA~\cite{wang2025opera}.
Given the full sequence of session actions, we predict the next user action
and evaluate along four dimensions following OPeRA: (1)~action generation
accuracy (exact match of the predicted action), (2)~action type F1 over
high-level categories (click, scroll, navigation, input, tab),
(3)~click type F1 over specific click subtypes, and (4)~session outcome
prediction (purchase vs.\ terminate).
We compare three persona conditions: (i)~\emph{GT persona}, the ground-truth
persona assembled from OPeRA's interview transcripts, demographics, personality
profiles, and shopping rationale; (ii)~\emph{Multi-hyp persona}, our
IToM-inferred persona (best-scored hypothesis from Stage~5); and
(iii)~\emph{No persona}, a control with no user context.
All conditions use the same session prefix and prompt template provided by the
OPeRA paper.
We evaluate on two LLM backends: Claude~Sonnet~3.7 (used in the original OPeRA
evaluation) and Claude~Opus~4.6.

\begin{table}[!tbh]
\centering
\caption{Next action prediction (15~users, 90~sessions, N=985
instances). Act.~Gen.: exact-match accuracy of the full predicted action
string; AT Wt\,/\,Mac-F1: weighted and macro F1 over high-level action
categories (click, scroll, input, etc.); Click Wt-F1: weighted F1 over
click subtypes; SO Acc\,/\,Wt-F1: accuracy and weighted F1 for
predicting session outcome (purchase vs.\ terminate).
Best results per-model in \textbf{bold}.}
\tablevspace
\label{tab:action_prediction}
\resizebox{\columnwidth}{!}{%
\begin{tabular}{llcccccc}
\toprule
\textbf{Model} & \textbf{Persona} & \textbf{Act.\ Gen.$\uparrow$} & \textbf{AT Wt-F1$\uparrow$} & \textbf{AT Mac-F1$\uparrow$} & \textbf{Click Wt-F1$\uparrow$} & \textbf{SO Acc$\uparrow$} & \textbf{SO Wt-F1$\uparrow$} \\
\midrule
\multirow{3}{*}{Sonnet 3.7}
  & No persona  &  9.14 & 84.78 & 32.28 & \textbf{47.07} & 62.22 & 54.67 \\
  & GT persona  &  8.83 & \textbf{86.18} & \textbf{35.53} & 47.00 & 60.00 & 56.47 \\
  & Multi-hyp (ours)   & \textbf{11.78} & 83.26 & 29.62 & 46.59 & \textbf{65.56} & \textbf{60.06} \\
\midrule
\multirow{3}{*}{Opus 4.6}
  & No persona  & 19.09 & 80.21 & 30.66 & 47.05 & 64.44 & 61.87 \\
  & GT persona  & 17.06 & 78.20 & 29.98 & 43.16 & 65.56 & 62.80 \\
  & Multi-hyp (ours)   & \textbf{25.69} & \textbf{83.90} & \textbf{32.36} & \textbf{49.78} & \textbf{68.89} & \textbf{68.35} \\
\bottomrule
\end{tabular}%
}
\end{table}

\paragraph{Results.}
Table~\ref{tab:action_prediction} presents the results.
On Claude~Opus~4.6, our inferred persona substantially outperforms the
ground-truth persona on action generation accuracy (25.69\% vs.\ 17.06\%,
$+$50.6\% relative), action type weighted F1 (83.90\% vs.\ 78.20\%), and click
type weighted F1 (49.78\% vs.\ 43.16\%).
Both persona conditions outperform the no-persona baseline on action generation
and action type F1, confirming that persona conditioning provides meaningful
signal.
On Claude~Sonnet~3.7, our persona achieves higher action generation accuracy
(11.78\% vs.\ 8.83\%) and session outcome accuracy (65.56\% vs.\ 60.00\%)
while showing slightly lower action type F1.
Our Sonnet GT~persona results are consistent with OPeRA's published numbers,
validating our reproduction.

OPeRA's own analysis noted that persona conditioning does not consistently
improve action generation accuracy and may introduce noise into step-level
reasoning, even if it helps action type and click type
classification~\cite{wang2025opera}.
Our belief-grounded personas partially mitigate this issue: on
Claude~Opus~4.6, the inferred persona improves action generation ($+$50.6\%
relative over GT) \emph{and} classification metrics simultaneously, suggesting
that grounding personas in specific behavioral episodes rather than abstract
trait summaries reduces the noise problem OPeRA identified.

\subsection{RQ2: Shopping Category Prediction}
\label{sec:rq2}

\paragraph{Setup.}
To evaluate downstream predictive power, we test whether the persona can
predict which product categories a user will shop for in held-out sessions.
We use a 70/30 session holdout split: personas are constructed from 70\% of
each user's sessions, and the remaining 30\% provide
ground truth.
Ground truth categories are extracted from held-out search terms and product
interactions, with LLM-based reclassification into 11~canonical categories
(e.g., Health \& Personal Care, Electronics \& Technology, Home \& Kitchen).
This yields 27~users with non-empty holdout ground truth (mean ${\sim}$4
categories per user).

\begin{table}[!t]
\centering
\caption{Shopping category prediction (N=27). Best in \textbf{bold}.}
\tablevspace
\label{tab:category_baselines}
{\footnotesize\setlength{\tabcolsep}{4pt}\renewcommand{\arraystretch}{0.95}
\resizebox{\linewidth}{!}{%
\begin{tabular}{lccccc}
\toprule
\textbf{Method} & \textbf{Mac-P$\uparrow$} & \textbf{Mac-R$\uparrow$} & \textbf{Mac-F1$\uparrow$} & \textbf{NDCG@5$\uparrow$} & \textbf{Hit@3$\uparrow$} \\
\midrule
Popularity           & .266 & 1.00 & .409 & .581 & 88.9 \\
User History         & .425 & .689  & .504 & .615 & 85.2 \\
Recency-Weighted     & .371 & .753  & .458 & .615 & 81.5 \\
Zero-shot LLM~\cite{hou2024large}
                     & \textbf{.559} & .500  & .498 & .522 & 88.9 \\
Structured Summary~\cite{christakopoulou2023interest}
                     & .533 & .689  & .562 & .636 & 88.9 \\
CoT Prompting~\cite{wei2022chain}
                     & .501 & .730  & .567 & .658 & 88.9 \\
\midrule
\textbf{Multi-hyp (ours)} & .499 & \textbf{.764} & \textbf{.567} & \textbf{.701} & \textbf{100} \\
\bottomrule
\end{tabular}}}
\end{table}

\paragraph{Baselines.}
We compare against two families of baselines
(Table~\ref{tab:category_baselines}).
\emph{Heuristic} baselines operate on category co-occurrence statistics from
training sessions: Popularity predicts globally frequent categories, User
History echoes per-user training categories ranked by frequency, and
Recency-Weighted applies exponential decay to favor recent sessions.
\emph{LLM-based} baselines use the same backbone model (Claude~Opus~4.6) and
identical raw session data, isolating the contribution of our belief-inference
and persona-synthesis stages: 1) Zero-shot LLM directly prompts the model to
predict categories from raw action
logs~\cite{hou2024large}, 2) Structured Summary first compresses
sessions into browsing summaries before
predicting~\cite{christakopoulou2023interest}, and 3) CoT Prompting adds
explicit step-by-step reasoning over browsing
patterns~\cite{wei2022chain}.
We omit neural sequential models~\cite{kang2018self, sun2019bert4rec, hidasi2016session}, as they require item-ID interaction sequences and orders of magnitude more users for training. OPeRA provides
27~users with atomic browsing traces (clicks, scrolls, typed queries) rather than item-level interaction logs, making these architecturally inapplicable.

\paragraph{Results.}
Table~\ref{tab:category_baselines} presents the comparison.
Among heuristic baselines, Popularity achieves perfect recall by predicting
all 11~categories but at low precision (.266), while User History obtains the best heuristic F1 (.504) through direct category overlap.
Among LLM baselines, adding prompt structure could possibly improve results:
Zero-shot LLM achieves the highest precision (.559) but lowest recall (.500),
and CoT Prompting reaches the highest baseline F1 (.567) and recall (.730). Our multi-hypothesis persona matches the best baseline F1 (.567) and
achieves NDCG@5 of .701 ($+$6.5\% over CoT) with Hit@3 of 100\%.

\subsection{RQ3: Big Five Personality Prediction}
\label{sec:rq3}

\paragraph{Setup.}
OPeRA provides self-reported Big Five scores on a 1--5 scale (Extremely Low to
Extremely High) for 46~users.
Each inference mode predicts Big Five trait scores from the persona;
predictions are compared against ground truth via MAE and Pearson~$r$.

\paragraph{Baselines.}
We compare against:
\emph{Random Uniform} (randomly guess 1--5 per trait),
\emph{Scale Midpoint} (always predict 3.0),
\emph{Population Norms} (published Big Five means: EX=3.2, AG=3.6, CO=3.4,
ES=3.1, IN=3.4), and
\emph{GT Mean} (per-trait mean of all 46 users, an oracle baseline requiring
access to the ground truth distribution).

\begin{figure}[!b]
\centering
\includegraphics[width=\columnwidth]{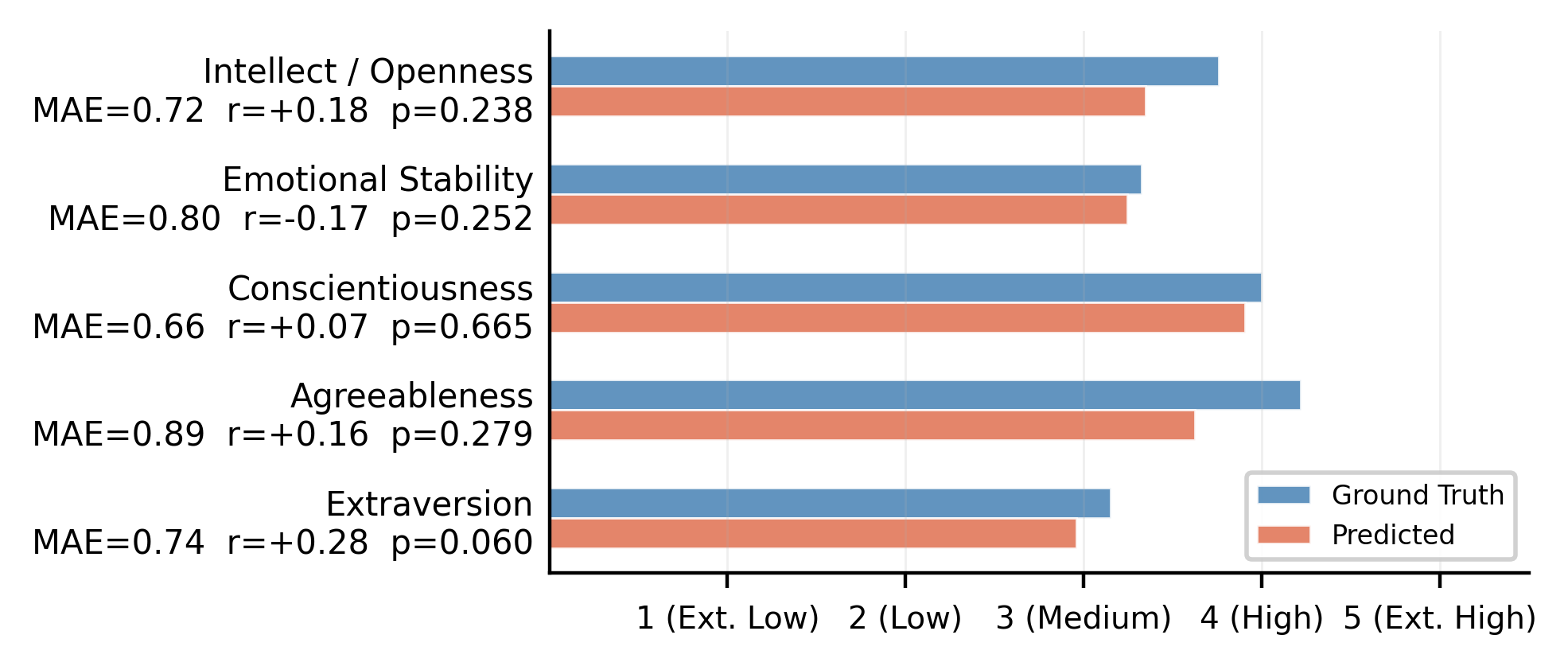}
\caption{Big Five personality prediction results (N=46). Mean trait scores comparing self-reported ground truth (survey) against our multi-hypothesis IToM predictions. Per-trait MAE, Pearson correlation ($r$), and $p$-values are shown alongside each trait.}
\Description{Grouped bar chart comparing self-reported Big Five trait scores (ground-truth survey) against our multi-hypothesis IToM predictions across 46 users, for the five traits: Openness, Conscientiousness, Extraversion, Agreeableness, and Emotional Stability. For each trait, two bars show the mean ground-truth score and the mean predicted score on a 1-to-5 scale, with per-trait mean absolute error, Pearson correlation $r$, and $p$-value annotated next to the bar pair.}
\label{fig:big5_bars}
\end{figure}

\paragraph{Results.}
As shown in Table~\ref{tab:baselines_summary} and
Figure~\ref{fig:big5_bars}, our method achieves an overall MAE of 0.762,
outperforming all non-oracle baselines including population norms (0.836,
$-$8.8\% relative improvement) and closing 88\% of the gap from random
guessing to the oracle mean.
Four of five traits show positive Pearson correlations with ground truth,
with Extraversion approaching significance ($r{=}+0.280$, $p{=}0.060$).

Multi-hypothesis aggregation shifts the aggregate correlation into the positive range on this fixed cohort (bootstrap 95\% CI $[-0.05,\,+0.26]$; $N{=}46$, dataset-given), reversing the negative aggregate of single-interpretation baselines. We therefore scope RQ3 as a \emph{population-level} claim about the direction of LLM persona bias and its correction; per-trait tests are underpowered at this sample size.


\begin{table}[!t]
\centering
\caption{Big Five MAE comparison (N=46). Our multi-hypothesis method
outperforms all non-oracle baselines.}
\tablevspace
\label{tab:baselines_summary}
{\footnotesize\setlength{\tabcolsep}{4pt}\renewcommand{\arraystretch}{0.9}
\begin{tabular}{lc}
\toprule
\textbf{Method} & \textbf{Avg MAE$\downarrow$} \\
\midrule
Random Uniform (1--5)         & 1.495 \\
Scale Midpoint (predict 3.0)  & 1.022 \\
Direct Inference              & 0.945 \\
Questionnaire (90 items)      & 0.945 \\
Population Norms              & 0.836 \\
\textbf{Multi-Hypothesis (M=7)} & \textbf{0.762} \\
\midrule
GT Mean (oracle)              & 0.662 \\
\bottomrule
\end{tabular}}
\end{table}

\paragraph{Ablation: Why multi-hypothesis reasoning matters.}
A key question is whether the multi-hypothesis aggregation contributes beyond
what simpler inference strategies achieve.
We compare against two established approaches from the LLM personality
literature, applied to the \emph{same} persona profiles:
(i)~\emph{Direct inference}, where the LLM reads a single persona and directly
estimates trait scores~\cite{wang2025emulate,huang2026designing}; and
(ii)~\emph{Questionnaire administration}, where the LLM role-plays as the
persona and answers standardized personality items
individually~\cite{dewinter2024chatgpt,pellert2024ai,bhandari2025evaluating}.
Both approaches have been shown to produce reasonable personality profiles when
LLMs role-play from explicit trait
descriptions~\cite{wang2025emulate,park2024generative}; here we test them on
behaviorally-inferred personas where trait levels are \emph{not} given.

Table~\ref{tab:mode_ablation} shows our results that single-interpretation methods produce \emph{negative} average Pearson correlations with ground truth ($-$0.103 for questionnaire, $-$0.008 for direct inference), meaning their predictions are \emph{anti-correlated} with actual personality.
Both methods achieve MAE$=$0.945, worse than simply predicting published
population norms for every user (0.836).
Our multi-hypothesis method reverses the average correlation from $-0.103$ (Direct Inference) and $-0.008$ (Questionnaire) to $+0.103$, a qualitative shift from systematically wrong to directionally correct, and cuts MAE from $0.945$ (both unaided-LLM baselines) to $0.762$. Population Norms ($0.836$) is an informed prior of published Big Five means, not an unaided-LLM comparison. See Figure~\ref{fig:big5_bars} for per-trait statistics between our method and ground-truth.


\begin{table}[!t]
\centering
\caption{Ablation: inference method comparison (N=46). All methods use
identical persona profiles. Single-interpretation baselines from the LLM
personality literature produce anti-correlated predictions; multi-hypothesis
aggregation reverses the sign.}
\tablevspace
\label{tab:mode_ablation}
{\footnotesize\setlength{\tabcolsep}{4pt}\renewcommand{\arraystretch}{0.9}
\begin{tabular}{lcc}
\toprule
\textbf{Method} & \textbf{Avg MAE$\downarrow$} & \textbf{Avg $r$$\uparrow$} \\
\midrule
Questionnaire~\cite{dewinter2024chatgpt} & 0.945 & $-$0.103 \\
Direct Inference~\cite{wang2025emulate}  & 0.945 & $-$0.008 \\
\textbf{Ours: Multi-Hyp.\ (M=7)} & \textbf{0.762} & \textbf{+0.103} \\
\bottomrule
\end{tabular}}
\end{table}

We term this negative correlation as \emph{persona bias}: the LLM maps ambiguous behavioral
evidence to stereotypical personality profiles, producing overconfident
predictions that are \emph{worse than knowing nothing}.
We analyze this phenomenon further in the Discussion.

\subsection{RQ4: Shopping Preference Survey Alignment}
\label{sec:rq4}

\paragraph{Setup.}
OPeRA includes 12~Likert shopping attitude items per user (1=Strongly Disagree
to 5=Strongly Agree), covering research habits, price sensitivity, brand
preference, and decision-making style.
We treat these as a structured questionnaire: the LLM infers how the persona
would respond to each item, and predicted responses are compared numerically
against ground truth.
Per-item accuracy is computed as $\mathrm{acc}_i = 1 - |y_i -
\hat{y}_i| / 4$, normalizing Likert distance to $[0, 1]$.

\paragraph{Results.}
Our method achieves 76.6\% normalized accuracy (MAE = 0.937 on a 1--5 Likert scale, N$=$46), outperforming both single-interpretation ablation baselines:
direct inference (71.9\%) and LLM-administered questionnaire
(72.5\%).%
\footnote{Direct inference asks the LLM to predict all survey responses from
the best persona hypothesis in a single call; the questionnaire mode
administers items individually with the LLM role-playing as the
persona~\cite{dewinter2024chatgpt,pellert2024ai}.}

\begin{figure}[bh]
\centering
\includegraphics[width=\columnwidth]{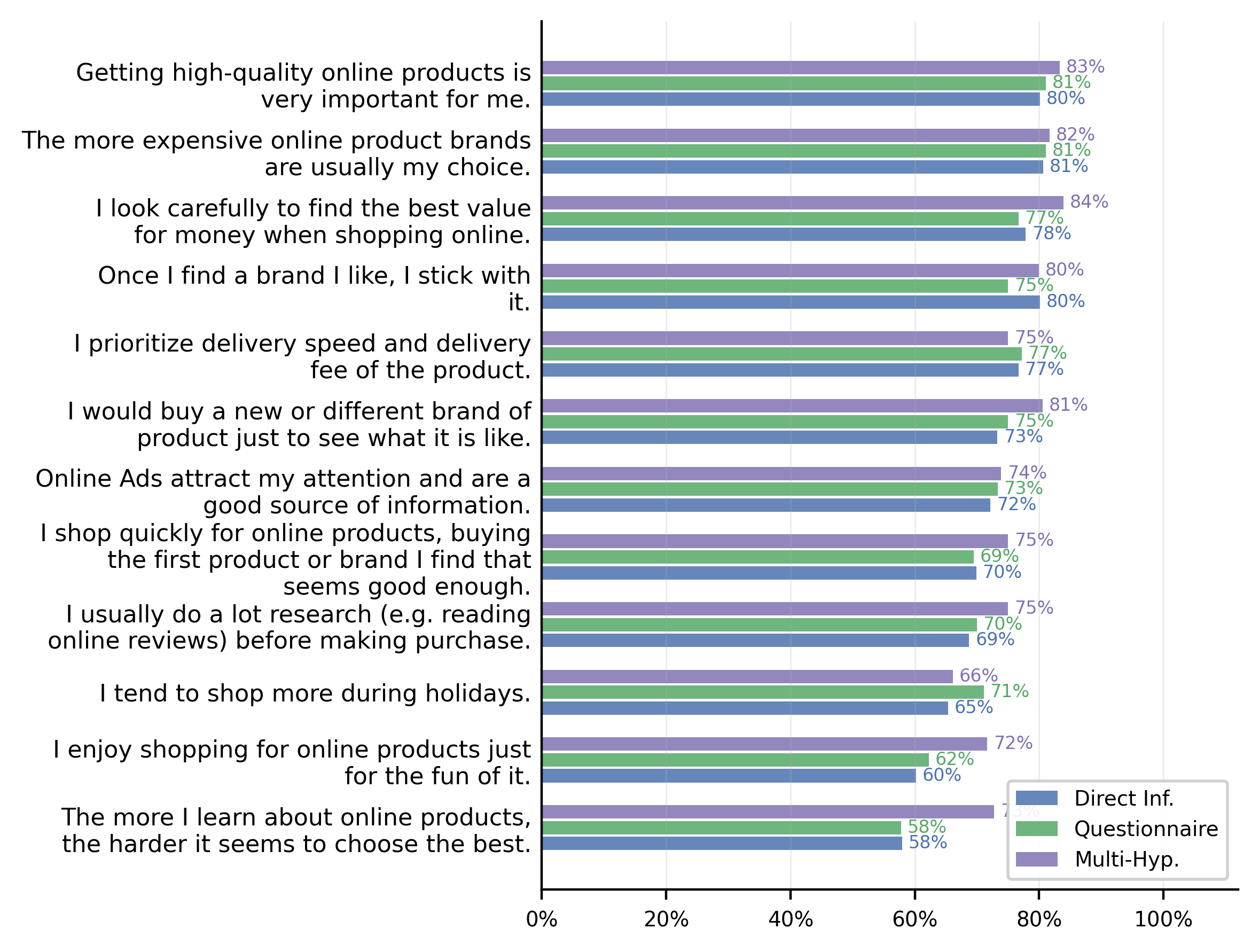}
\caption{Per-item shopping attitude accuracy (N=46). Our multi-hypothesis
method achieves the highest accuracy on 10 of 12 items compared to
single-interpretation ablations.}
\label{fig:shopping_survey}
\Description{Grouped horizontal bar chart showing per-item shopping attitude
accuracy for direct inference, questionnaire, and multi-hypothesis modes.}
\end{figure}

Figure~\ref{fig:shopping_survey} shows per-item accuracy.
Several items (e.g., ``I usually do a lot of research before making a
purchase'') exceed 0.85 accuracy, demonstrating that the persona captures
genuine shopping attitude signal across diverse dimensions, from research
habits to price sensitivity to decision confidence.

\subsection{Discussion}

\paragraph{Inferred personas match or exceed ground-truth personas.}
The action prediction results (Table~\ref{tab:action_prediction}) show that personas inferred purely from behavioral traces achieve competitive or higher scores than ground-truth personas assembled from interviews, demographics, and personality assessments on most prediction metrics ($+$50.6\% relative action generation accuracy on Claude~Opus~4.6). This advantage reflects a representational-format difference as much as a user-modeling depth difference: GT personas are abstract trait summaries that anchor the model to a fixed characterization~\cite{gupta2024persona}, while our belief-grounded personas encode specific decision episodes that translate more directly into step-level prediction without imposing a rigid role frame. This result should be read as evidence that our episode-encoding format is usable at inference time, not that inferred personas understand the user better than GT personas in an absolute sense: the $+$50.6\% margin may partly reflect prompt-format alignment with the action-prediction task, and a format-controlled comparison (rewriting GT personas into episode style, ablating episodic content from inferred personas) is left to future work. The value of the inferred pipeline is not that it beats GT in absolute score, but that it (a)~removes the qualitative-data collection bottleneck (interviews, demographics, personality assessments) that makes GT personas impractical in production, and (b)~yields an interpretable, auditable natural-language artifact supporting downstream tasks that raw click-stream data cannot.

\paragraph{Persona bias and multi-hypothesis reasoning.}
The ablation (Table~\ref{tab:mode_ablation}) reveals what we term
\emph{persona bias}: single-interpretation methods map ambiguous behavioral
evidence to stereotypical personality profiles (e.g., interpreting every
careful shopper as an anxious introvert), producing predictions
\emph{anti-correlated} with ground truth~\cite{bhandari2025evaluating,huang2026designing}.
This bias is especially severe for behaviorally inferred personas where
trait levels are uncertain.
Multi-hypothesis reasoning reverses the correlation sign (negative $\to$
positive) by generating $M$~diverse hypotheses with per-trait evidence
confidence and adaptively weighting toward population priors when evidence
is weak, providing a ``floor guarantee'' against persona bias.
The underlying mechanism is the \emph{situational} nature of
online shopping: users settle for ``good enough'' options under time pressure, get distracted by promotions, and make choices they might not replicate an hour
later due to situational variability and mood swings~\cite{simon1955behavioral}.
A single interpretation forces one coherent story onto inherently noisy
actions; multiple hypotheses allow the weight of evidence to point in
roughly the right direction without requiring every single hypothesis to be entirely
correct.

\paragraph{The role and limits of personality in behavioral modeling.}
Our results suggest that personality, as captured by standardized instruments
like the Big Five, may be more useful for explaining broad attitudinal
tendencies than for predicting specific actions.
The shopping preference survey (76.6\% accuracy) aligns well with inferred
personas, likely because survey items ask about general dispositions that
map naturally onto personality dimensions.
In contrast, traits like Emotional Stability are systematically misread from
browsing cues, and even well-calibrated traits (e.g., Conscientiousness)
show compressed prediction variance, suggesting the pipeline captures
population-level tendencies better than individual extremes.
This is perhaps unsurprising: a user's decision to compare three products
before purchasing reflects not only trait-level conscientiousness but also
situational factors, domain expertise, price sensitivity, and personal values
that personality inventories were never designed to
capture~\cite{vazire2008knowing}.
Personality may thus serve better as a \emph{soft prior} that anchors
attitudinal inference than as a direct predictor of behavior, where the
richer, episode-grounded belief representation carries more of the
explanatory weight.

\section{Case Study: Persona-Driven Spatial Interface}
\label{sec:case_study}

The experiments above show that IToM-inferred personas capture meaningful behavioral and attitudinal signal. We now present a case study showing how the same persona expressed in natural language can drive a different modality. We illustrate this transferability with a proof-of-concept spatial banking application built on Apple~VisionOS due to the substantial discrepancy between traditional web interfaces and spatial user interfaces.

\begin{figure}
\centering
\includegraphics[width=\columnwidth]{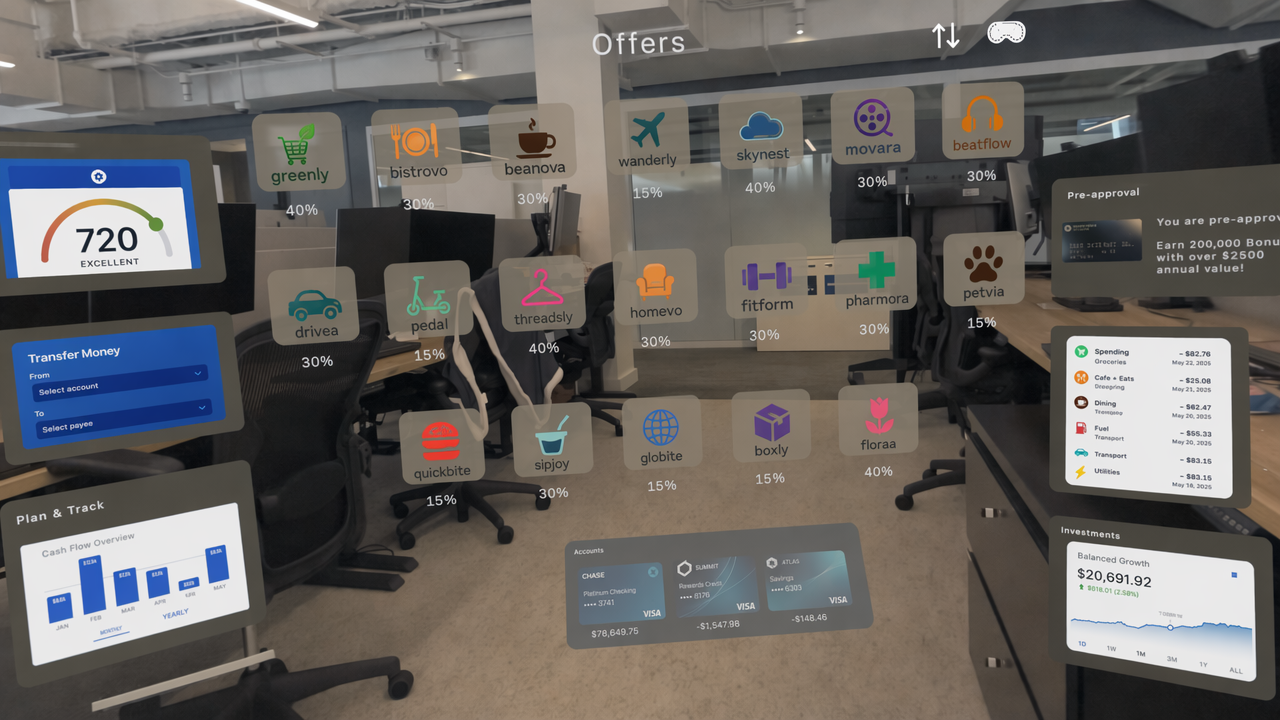}
\caption{Persona-driven spatial banking prototype on VisionOS. The left and right panels contain different
widgets placed in different positions and order, while the center stage
surfaces different offers, all determined by the user's inferred persona constructed from 2D browsing data (All logos and brand names shown are fictitious and for illustrative purposes only. Any resemblance to real trademarks or companies is coincidental; no affiliation or endorsement is implied.)} 
\Description{A screenshot of a persona-driven spatial banking application on Apple VisionOS, showing a multi-panel 3D interface. The interface presents common banking widgets, credit score, spending details, savings, together with persona-adapted content including merchant offers and recommended banking products. }
\label{fig:xr_case_study}
\end{figure}

\textit{Application.}
The prototype is a spatial banking app whose layout is fully persona-driven:
each user sees a structurally different interface
(Figure~\ref{fig:xr_case_study}).
The interface comprises three spatial panels, each defined by its
\emph{function} rather than fixed content:
a \emph{center stage} for primary tasks (account management, shopping offers);
a \emph{left panel} for financial health tools (credit journey, budgeting,
transfers); and
a \emph{right panel} for product discovery (credit cards, transaction
history).
The persona determines \emph{what} appears in each
panel and \emph{how} they are arranged
(e.g., a budget-conscious user sees spending breakdowns and savings trackers
prominently, while an investment-oriented user sees portfolio summaries and
market alerts).
Users can also rearrange, add, or dismiss widgets---interactions that feed
back into belief inference, iteratively refining the persona and subsequent layout adaptations.

\textit{Persona as cross-modal bridge.}
Shopping category predictions (RQ4) map directly to center-stage offer selection, while broader persona fields (life context, values, motivational
drives) inform which financial tools and products to surface and how prominently to position them.
The persona is used as natural-language context in a task-specific prompt, following the same pattern validated in our experiments.
This case study implementation serves as initial evidence of two transfer axes that interaction-based methods cannot easily address:
\emph{modality transfer}, where beliefs from 2D click streams inform 3D spatial layout; and
\emph{task transfer}, where shopping browsing behavior drives financial product suggestions and widget prioritization, a distinct application with no shared behavioral signal yet informed by the persona's cognitive profile. This opens opportunities to address cold-start challenges in emerging application domains where data is scarce.

\section{Conclusion}

We presented an Inverse Theory of Mind pipeline that reasons backward from
observed interaction traces to infer the beliefs, intentions, and
decision-making traits that explain user behavior, producing structured
natural-language personas that capture not only \emph{what} a user prefers but
\emph{who} they are.
Evaluated on 46 OPeRA users, the pipeline achieves 76.6\% shopping attitude
accuracy, reduces Big Five MAE below an informed population-norms prior ($0.762$ vs.\ $0.836$), corrects the sign of the average trait correlation relative to unaided-LLM baselines, and
predicts held-out shopping categories with Hit@3$=$100\%.
Without our multi-hypothesis reasoning, single-interpretation baselines
produce predictions \emph{anti-correlated} with ground truth, while
aggregating across diverse persona hypotheses reverses the correlation sign.
Because the resulting personas are expressed in natural language rather than
interface-specific embeddings, they transfer directly to generative UI engines
or spatial XR interfaces without retraining, supporting content
\emph{creation}, not just selection.

\textit{Limitations and future work.}
Evaluation is bounded by a single domain (Amazon shopping) and moderate sample
size ($N{=}46$ for personality).
Prediction variance is compressed, indicating well-calibrated means but limited differentiation at user extremes.
Emotional Stability is anti-predicted due to systematic LLM bias that adaptive
priors only partially mitigate.
The XR case study is illustrative; 
a controlled study comparing persona-driven layouts against a non-personalized baseline on task completion, subjective preference, and offer relevance is left to future work.
Future work will extend to multi-domain traces, larger populations, end-to-end
personalization in spatial and generative interfaces, and calibration methods
for improved per-user differentiation. These limitations largely stem from the limited availability of public datasets that provide rich interaction traces, environmental context, and user traits. Our work sheds light on user modeling through the IToM framework and opens new opportunities for future research on data-efficient, generalizable, and personalized interactive systems.

\section*{Disclaimer}
This paper was prepared for informational purposes by the Global Technology Applied Research and Consumer \& Community Banking of JPMorgan Chase \& Co. This paper is not a product of the Research Department of JPMorgan Chase \& Co. or its affiliates. Neither JPMorgan Chase \& Co. nor any of its affiliates makes any explicit or implied representation or warranty and none of them accept any liability in connection with this paper, including, without limitation, with respect to the completeness, accuracy, or reliability of the information contained herein and the potential legal, compliance, tax, or accounting effects thereof. This document is not intended as investment research or investment advice, or as a recommendation, offer, or solicitation for the purchase or sale of any security, financial instrument, financial product or service, or to be used in any way for evaluating the merits of participating in any transaction.

\bibliographystyle{ACM-Reference-Format}
\bibliography{reference}



\end{document}